\documentclass[11pt]{article}

\usepackage[final]{acl}

\usepackage{times}
\usepackage{latexsym}

\usepackage{amsmath} 
\usepackage[T1]{fontenc}
\usepackage{stfloats}
\usepackage[utf8]{inputenc}
\usepackage{float}

\usepackage{microtype}

\usepackage{inconsolata}

\usepackage{graphicx}
\usepackage{hyperref}
\title{Frame In, Frame Out: \\Measuring Framing Bias in LLM-Generated News Summaries}

\author{Valeria Pastorino \and Nafise Sadat Moosavi \\
  Department of Computer Science \\ University of Sheffield (UK) \\ 
  \texttt{ \{vpastorino1|n.s.moosavi\}@sheffield.ac.uk}}

\begin{document}
\maketitle
\begin{abstract}
News headlines and summaries shape how events are interpreted through selective emphasis and omission, a phenomenon commonly referred to as framing. Large language models are now routinely used to generate such content, yet existing evaluation frameworks largely overlook this dimension. We introduce Frame In, Frame Out (FIFO), the first large-scale benchmark for measuring framing presence in LLM-generated news summaries, grounded in the widely used XSum dataset. FIFO combines 15,499 jury-annotated examples with 320 expert-labeled instances ($\kappa = 0.61$) to validate and calibrate model-based annotations. Using FIFO, we analyze measured framing rates across 27 summarization models. We find that LLM-generated summaries often exhibit higher calibrated framing rates than human-written references, with substantial variation across topics and training regimes, including elevated rates in scientific and public health summaries. Our results establish framing as an underexplored and consequential dimension of summarization quality.
\end{abstract}

\section{Introduction}
 Framing is a communication strategy through which the interpretation of events can be shaped by selecting, emphasizing, omitting, and organizing information in particular ways \cite{entman_framing_1993,goffman1974frame}. From a semantic perspective, framing concerns meaning beyond propositional content: how lexical and discourse choices guide interpretation of the same underlying facts. In media and political communication, it influences attribution of responsibility \cite{iyengar1994anyone}, political tolerance \cite{nelson1997media}, and public opinion. As natural language generation systems are increasingly used to produce headlines, news briefs, and content summaries \cite{sadeghi2023newsbots}, they inherit a growing role in shaping public discourse. 

Despite extensive study in the social sciences, framing remains largely absent from the evaluation of model-generated text. 
In our setting, this absence is not merely theoretical: a system-generated summary can introduce an interpretive lens that is absent from the human-written reference summary for the same article.

\begin{quote}
\textbf{Gold summary (\texttt{Not Framed}):} ``Donkey, water buffalo and goat meat have been sold as burgers and sausages in South Africa, a study says.''\\
\textbf{System summary (\texttt{Framed}):} ``South Africa has been hit by a growing scandal over the sale of meat products that contain animals''
\end{quote}

In these examples from our data, both statements can be compatible with the underlying facts, but the system summary foregrounds a more evaluative interpretation of the event, steering the reader toward a particular understanding of its significance. This kind of perspective shift is largely invisible to standard summarization metrics.
In fact, evaluating natural language generation systems is an ongoing and challenging problem, with existing frameworks primarily focusing on factuality \cite{pagnoni-etal-2021-understanding, huang2024ufounifiedflexibleframework}, coherence \cite{fabbri-etal-2021-summeval}, and coverage \cite{liu-etal-2023-revisiting}. These dimensions capture important aspects of quality, but do not account for how information is selectively emphasized or omitted in ways that may contribute to political agendas, polarization or distorted public discourse. As a result, a summary may be factually accurate and fluent, yet still convey a particular perspective in subtle and socially consequential ways \cite{Puccetti_2024}. 

We address this gap with \textbf{Frame In, Frame Out (FIFO)}, the first benchmark for evaluating framing bias in summarization. FIFO provides both expert-annotated gold labels and model-generated silver labels for thousands of summaries, along with a calibration protocol to produce expert-calibrated framing estimates. Building on work in framing detection, we propose framing as an evaluation dimension for summarization, enabling systematic model- and topic-level comparisons of generated outputs.

Using FIFO, we analyze 27 summarization models spanning architectures and fine-tuning regimes. We find that several large language models produce higher calibrated framing rates than human-written baselines, with substantial variation by model size, training setup, and topic. Notably, elevated framing emerges even in domains such as science and public health, where neutrality is typically expected. These findings establish framing as a measurable and consequential property of LLM-generated summaries, and motivate its inclusion in future evaluation frameworks.

\section{Related Work}

Framing has mainly been studied as a supervised detection task using annotated corpora such as the Media Frames Corpus \citep{card_media_2015} and GVFC \citep{liu_detecting_2019}. Earlier work relied on topic models and lexical cues \citep{DIMAGGIO2013570, burscher2016}, followed by transformer-based classifiers \citep{khanehzar-etal-2019-modeling, classifying_2017}.
More recent approaches use LLMs such as GPT-4 to classify or explain frames in news text \citep{pastorino2024decodingnewsnarrativescritical, maab-etal-2024-media, lin-etal-2024-indivec}. Across these works, framing is treated as detection: the goal is to identify which frame a text expresses, if any. Our study differs by treating framing as an evaluation dimension, asking whether summarization systems introduce unintended or asymmetric framing.

Summarization evaluation has largely centered on coherence, factuality, content selection, and human preference judgments \citep{fabbri-etal-2021-summeval, pagnoni-etal-2021-understanding, liu-etal-2023-revisiting, bhandari-etal-2020-evaluating}. However, despite framing being central to how readers perceive neutrality, no existing benchmark evaluates framing as a dimension of generated summaries. Our work fills this gap by systematically analyzing framing in system-generated summaries, offering a new perspective on summarization quality.

\section{FIFO: Framing Bias in Summarization}
Our goal is to develop a scalable and reliable procedure for estimating framing in generated news summaries. This process results in \textbf{FIFO} (Frame In, Frame Out), a dataset composed of 15,499 model-generated summaries labeled via an LLM jury and 320 gold-standard annotations from human experts.
 We build on the best prompting strategy evaluated by \citet{pastorino2024decodingnewsnarrativescritical} to  use LLMs to classify summaries as either \texttt{Framed} or \texttt{Not Framed}. We extend this approach in two key ways:
(1) we introduce a human-annotated gold set to validate and estimate the reliability of model predictions, and
(2) we estimate expert-calibrated reliability weights from the agreement between silver and gold labels, and use them to compute calibrated framing rates across models.
This approach supports scalable, quantitatively grounded analysis anchored in expert judgment.

\paragraph{Summaries and Model Set}
Our analysis is based on the XSum dataset \cite{narayan-etal-2018-dont}, a single-sentence summarization corpus composed of BBC News articles and human-written summaries. 
Its extreme compression setting requires models to make strong content-selection and emphasis decisions, making it particularly well suited for studying framing in generated summaries. Because each output compresses a full article into a short summary, differences in what is foregrounded, backgrounded, or omitted become especially consequential. Moreover, XSum is a widely adopted benchmark, enabling systematic comparison across models and training regimes.

We analyse summaries from 27 systems sourced from \citet{liu2024llmsnarcissisticevaluatorsego} and \citet{panickssery2024llmevaluatorsrecognizefavor}, comprising a total of 15,499 summaries. These systems span a diverse range of architectures and training setups, including encoder-decoder models (e.g., BART \cite{lewis-etal-2020-bart}, T5 \cite{JMLR:v21:20-074}, FLAN-T5 \cite{chung_scaling_2022}), decoder-only models (e.g., GPT-2 \cite{Radford2019LanguageMA}, GPT-3 \cite{brown2020languagemodelsfewshotlearners}, GPT-4 \cite{openai2023gpt4}, GPT-Neo \cite{kashyap2023gptneocommonsensereasoning}, Claude \cite{TheC3}, Cohere \cite{cohere2024command}, LLaMA \cite{touvron2023llama2}). Fine-tuned variants are labeled with \textsc{-xsum} or \textsc{-cnn}, reflecting supervised training on XSum or CNN/DailyMail \cite{NIPS2015_afdec700}, respectively\footnote{All gold and silver framing annotations and model checkpoints (sourced from the original papers \cite{liu2024llmsnarcissisticevaluatorsego, panickssery2024llmevaluatorsrecognizefavor}) are available at \url{https://github.com/vpastorino/FIFO}}.

\paragraph{Framing Operationalization}
We annotate framing as a binary property of the generated summary. A summary is labeled \texttt{Framed} when it presents the article through an identifiable interpretive lens, including through selective emphasis or omission, evaluative wording, causal or moral attribution, responsibility assignment, or other discourse choices that make one interpretation more salient. A summary is labeled \texttt{Not Framed} when it conveys the core event without introducing such an interpretive lens. This binary formulation supports the benchmark’s primary goal: estimating how often summarization systems produce framed outputs. Fine-grained frame taxonomies are useful for analyzing which type of frame is present, but the presence of framing is the necessary first-order question for system-level evaluation.

\paragraph{Silver Labeling via LLM Jury}
To assign framing labels at scale, we apply the best-performing prompt from \citet{pastorino2024decodingnewsnarrativescritical} to a three-member LLM jury comprising GPT-4.1-nano, GPT-4o, and GPT-3.5-Turbo. Each model independently labels a summary as either \texttt{Framed} or \texttt{Not Framed}, and the final label is assigned via majority vote. The resulting labels constitute our silver dataset, which is used for large-scale analysis of framing across models.

\paragraph{Expert Annotations and Jury Validation}
To evaluate the reliability of the silver labels, we randomly sampled 320 model-generated summaries and annotated them for framing bias. Annotation was performed by a domain expert in linguistics and manual framing analysis, using informed judgment grounded in established framing theory \cite{entman_framing_1993, goffman1974frame}. Any uncertainties were resolved with a second expert, and labels were assigned by agreement.
We compared the resulting expert labels to the majority-vote predictions of the LLM jury, treating the expert annotations as the gold reference for validation. Agreement was measured using Cohen’s $\kappa$ \cite{cohen1960coefficient}, resulting in a score of 0.616, which falls within the range of substantial agreement \cite{landis1977measurement}. This level of alignment indicates that the silver labels produced by the LLM ensemble capture framing-relevant distinctions with reasonable reliability. 

\paragraph{Expert-Calibrated Reweighting}
\label{silver}
To adjust for prediction error in the silver labels, we estimate how often each jury label aligns with expert judgments using the 320 expert-annotated examples. Among items labeled \textit{Framed} by the jury, 77.8\% were also labeled \textit{Framed} by experts. Among items labeled \textit{Not Framed} by the jury, 16.3\% were nevertheless judged \textit{Framed} by experts. We assign each silver headline the corresponding calibrated value and estimate framing rates as the mean of these values rather than as the mean of raw binary jury labels. For a subset of summaries $S$, we define the expert-calibrated framing rate $\mathrm{FR}(S)$ as:

\[
\mathrm{FR}(S)=\frac{1}{|S|}\sum_{s\in S} w_s
\]

where $|S|$ denotes the number of summaries in $S$, and $w_s$ is the calibrated framing score assigned to summary $s$:

\[
w_s =
\begin{cases}
0.778 & \text{if } y_j(s)=F,\\
0.163 & \text{if } y_j(s)=N.
\end{cases}
\]

Here, $y_j(s)$ denotes the jury label for summary $s$, $F$ denotes \textit{Framed}, and $N$ denotes \textit{Not Framed}.

\section{Framing Behaviour Across Summarization Models}
\begin{figure}[ht]
    \centering
    \includegraphics[width=\columnwidth]{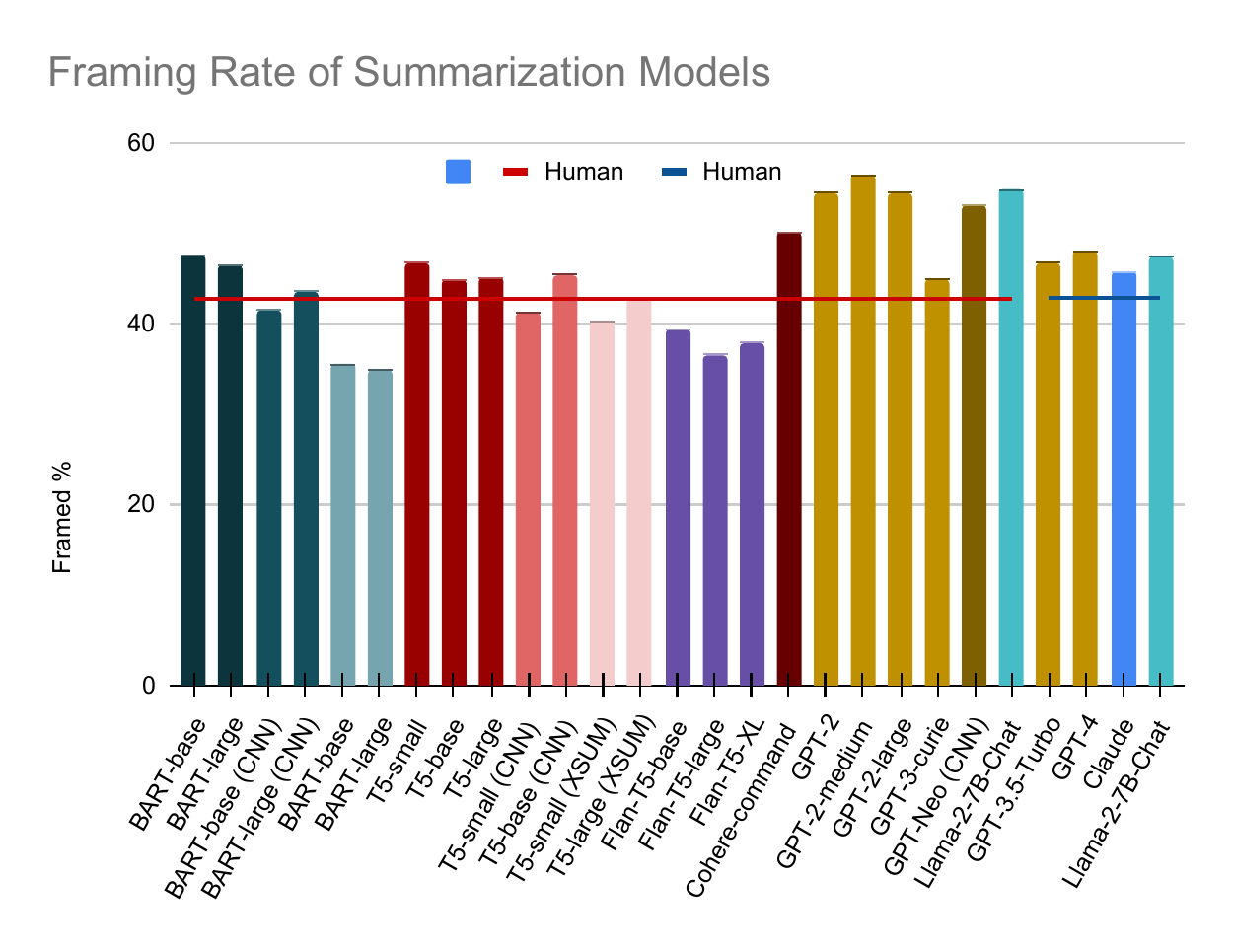} 
    \caption{Expert-calibrated framing rates across 27 summarization systems. Red and blue lines indicate human-authored baselines for the two subsets.}
    \label{fig:DEFmodels}
\end{figure}

The FIFO dataset enables systematic, model-level analysis of framing in summarization outputs. Figure~\ref{fig:DEFmodels} presents expert-calibrated framing rates across 27 systems\footnote{Unweighted results show similar trends but larger absolute differences and minor ranking shifts among mid-range models.}. 

These summaries are sourced from two prior model collections \cite{liu2024llmsnarcissisticevaluatorsego, panickssery2024llmevaluatorsrecognizefavor}, each associated with a distinct portion of the XSum dataset; human-authored reference summaries for each subset serve as framing baselines (indicated by red and blue horizontal lines).

\begin{figure*}[t]
\vspace{-0.5cm}
    \centering
    \includegraphics[width=1\linewidth]{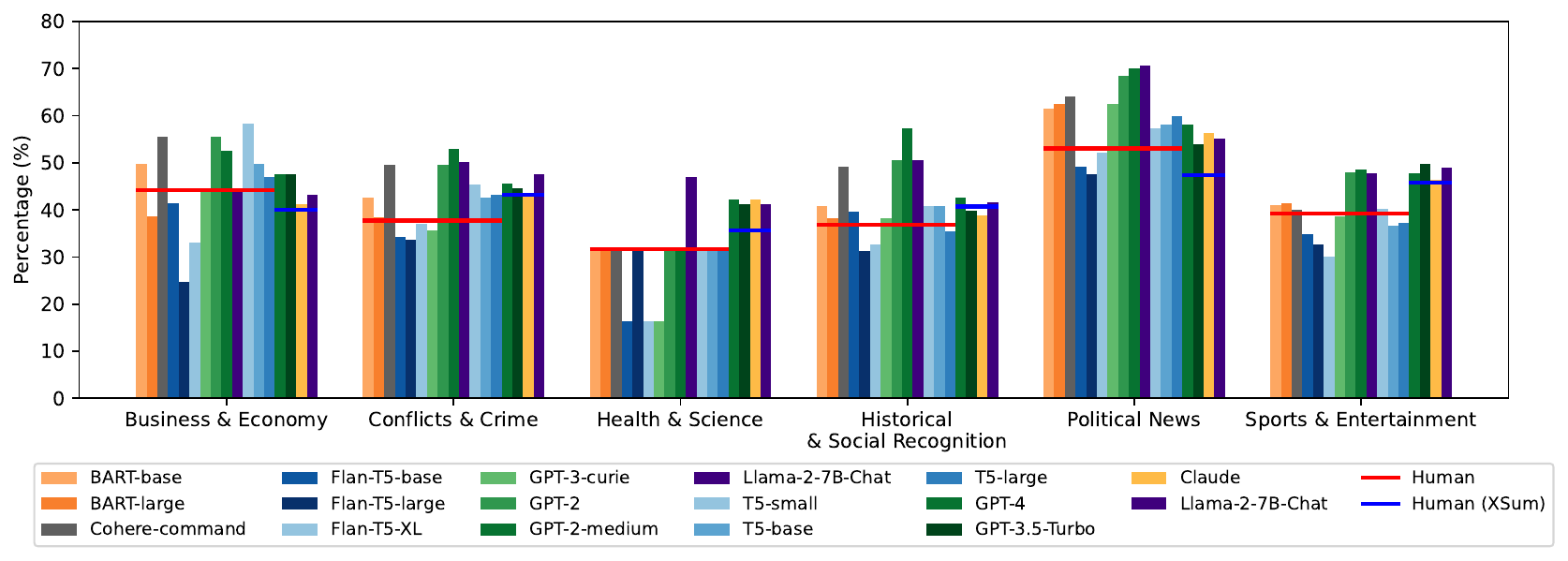}
    \caption{Framing rates by topic and model. Bars show the percentage of summaries labeled as \texttt{Framed}, grouped by topic. Each color corresponds to a model; horizontal lines indicate human framing baselines from two subsets.}
    \label{fig:topicsmodels}
\end{figure*}

\paragraph{Model Scale and Pretraining Scope} 
Smaller models (e.g., BART, FLAN-T5 variants) exhibit lower framing rates, often below the human baseline. However, this often reflects lower output quality: low quality summaries may lack the structure necessary for framing to be meaningfully expressed or detected. 
In contrast, larger models (e.g., GPT-4, Claude, LLaMA) produce more fluent and linguistically-rich summaries and correspondingly show higher framing rates. This difference in framing rates between smaller and larger models is statistically significant (paired t-test, p = 0.0012).

\paragraph{Effect of Fine-Tuning}
Models fine-tuned on XSum exhibit lower framing rates than their base counterparts. This reduction is statistically significant (p = 0.0006; 95\% CI: –19.27\% to –7.78\%) and suggests that task-specific finetuning is associated with lower framing rates in this setting.

\paragraph{Intra-Family Size Effects}
Within model families (e.g., BART, T5, GPT, FLAN-T5), we observe a moderate inverse correlation between model size and framing rate (Pearson $r = -0.44$), with larger variants tending to generate slightly less framed content. Together with the across-model trend above, this suggests that training regime and data may drive framing differences more strongly than parameter count alone.

These findings suggest that some high-capacity systems produce framing at rates above the corresponding human-authored baselines. This emphasizes the need for framing-aware evaluation in summarization, since standard quality metrics may miss interpretive shifts in otherwise fluent and factually compatible summaries.

\section{Framing as a Function of Topic}

To examine how framing varies across domains, we categorize each summary into one of six high-level topics: \textit{Business \& Economy}, \textit{Conflicts \& Crime}, \textit{Health \& Science}, \textit{Historical \& Social Recognition}, \textit{Political News}, and \textit{Sports \& Entertainment} using a hybrid process.
Framing rates are computed using the expert-calibrated reliability weights introduced in Section~\ref{silver}, ensuring that reported scores reflect validation against expert annotations.
Figure~\ref{fig:topicsmodels} shows that framing rates vary substantially by topic and model capacity. Human-authored summaries display clear topic sensitivity, with \textit{Health \& Science} exhibiting the lowest framing rate and \textit{Political News} the highest. Smaller models (e.g., BART, FLAN-T5 variants), which often produce lower-quality summaries, often fall below these human baselines across most topics. In contrast, larger, more capable models (e.g., GPT-4, Claude, LLaMA) frequently exceed human framing rates, particularly in \textit{Political News}, where nearly all high-capacity models surpass the already elevated human baseline ($\approx$53\%). Even in \textit{Health \& Science}, where humans frame only 31\% of instances, many larger models exceed this level. These findings show that framing in LLM-generated summaries is topic-dependent and often exceeds human norms in larger models, highlighting the need for framing-aware evaluation as such models are now widely used in large-scale content production, including news writing and headline generation \cite{Puccetti_2024}.

\paragraph{Length vs.\ Framing Presence}

To investigate the relationship between text length and framing, we computed a point-biserial correlation coefficient, which measures the association between a continuous variable (text length) and a dichotomous variable (framing status). Our analysis revealed a small yet statistically significant positive correlation, $r_{pb} \approx 0.1904$, indicating that longer texts are modestly associated with the presence of framing. Specifically, texts identified as \texttt{Framed} had an average length of 147 words, whereas those labeled \texttt{Not Framed} averaged 83 words. This indicates that length is associated with framing presence, but the small effect size suggests that framing is not reducible to length alone.

\section{Conclusions}
We introduced FIFO, a benchmark for evaluating framing in generated news summaries, combining expert-annotated gold labels with LLM-jury annotations that are reweighted using expert-calibrated reliability estimates. Across 27 systems, we found that several large language models produce higher framing rates than human-written baselines, with substantial variation by model capacity, training regime, and topic, including in domains such as health and science.
These results highlight a limitation of current evaluation frameworks, which prioritize accuracy and fluency but largely ignore how information is presented and interpreted. Our findings demonstrate that framing is a measurable and socially consequential dimension of summary quality that is not captured by existing metrics. As LLMs are increasingly deployed in content production, neglecting this dimension risks reinforcing polarized or distorted narratives.
By establishing framing as a tractable evaluation target, FIFO provides a foundation for incorporating perspective-sensitive analysis into future benchmarks and assessment practices.

\section{Limitations}
The findings of this study should be considered alongside some limitations. First, while our use of silver labels generated by a jury of large language models allows for broad coverage, these labels may reflect model-specific biases or blind spots. We mitigate this risk by validating the jury labels against a smaller expert-annotated gold set and using the observed expert–jury agreement to estimate aggregate framing rates.

Our study focuses on single-document summarization in English using the XSum dataset. This enables tight control over the input domain but leaves open questions about how framing manifests in more diverse linguistic and cultural settings.

Finally, our operationalization of framing is binary, distinguishing between \texttt{Framed} and \texttt{Not Framed} summaries. This choice matches FIFO’s goal of estimating whether framing is present in generated summaries and how often systems produce framed outputs. However, it necessarily simplifies a complex, context-dependent phenomenon and does not identify fine-grained frame types.

\section{Acknowledgements}

This work was supported by the Centre for Doctoral Training in Speech and Language Technologies (SLT) and their Applications funded by UK Research and Innovation [grant number EP/S023062/1]. We also acknowledge IT Services at The University of Sheffield for the provision of services for High Performance Computing.

\bibliography{custom}

\appendix

%\section{Example Appendix}
\label{sec:appendix}

\end{document}